\documentclass{article}

\PassOptionsToPackage{numbers,compress}{natbib}
\usepackage[preprint]{neurips_2026}

\usepackage[utf8]{inputenc} 
\usepackage[T1]{fontenc}    
\usepackage{hyperref}       
\usepackage{url}            
\usepackage{graphicx}
\usepackage{multirow} 
\usepackage{float}
\usepackage{subcaption}
\usepackage{booktabs}       
\usepackage{amsmath}
\usepackage{amsfonts}       
\usepackage{nicefrac}       
\usepackage{microtype}      
\usepackage[table]{xcolor}       

\title{RESBev: Making BEV Perception More Robust}

\author{%
Lifeng Zhuo \quad Kefan Jin \quad Zhe Liu \quad Hesheng Wang \\
Shanghai Jiao Tong University
}

\begin{document}

\maketitle

\begin{abstract}
  Bird's-eye-view (BEV) perception has emerged as a cornerstone of autonomous driving systems, providing a structured, ego-centric representation critical for downstream planning and control. However, real-world deployment faces challenges from sensor degradation and adversarial attacks, which can cause severe perceptual anomalies and ultimately compromise the safety of autonomous driving systems. To address this, we propose a resilient and plug-and-play BEV perception method (RESBev), which can be easily applied to existing BEV perception methods to enhance their robustness to diverse disturbances. Specifically, we reframe perception robustness as a latent semantic prediction problem. A latent dynamic predictor is constructed to extract spatiotemporal correlations across sequential BEV observations, thereby learning the underlying BEV state transitions to predict clean BEV features for reconstructing corrupted observations. The proposed framework operates at the semantic feature level of the BEV perception pipeline, enabling recovery that generalizes across both natural disturbances and adversarial attacks without modifying the underlying backbone. Extensive nuScenes experiments show RESBev notably enhances BEV perception robustness against natural disturbances and adversarial attacks, with latency, parameter and memory analyses verifying its excellent robustness-efficiency trade-off.
\end{abstract}

\section{Introduction}
\label{sec:intro}

The bird's-eye-view (BEV) representation has become a foundation of modern autonomous driving perception systems. By transforming multi-camera inputs into a unified top-down representation, it resolves projective ambiguities and provides geometrically consistent scene understanding essential for downstream tasks. Lift-Splat-Shoot (LSS)~\citep{philion2020lift} introduces a foundational framework for this transformation, pioneering camera-centric 3D perception. Building upon this paradigm, numerous BEV perception models have been proposed. For instance, BEVFusion~\citep{liu2022bevfusion} integrates camera features with LiDAR signals to improve geometric fidelity and robustness. Meanwhile, BEVFormer~\citep{li2022bevformer} leverages temporal self-attention to aggregate BEV features from historical frames, enabling the model to maintain persistent object tracks and handle ego-motion across frames.

Despite strong benchmark performance on datasets such as nuScenes~\citep{caesar2020nuscenes}, BEV models remain vulnerable to real-world corruptions and adversarial perturbations. 
Adverse conditions such as fog, darkness, snow, camera failures, and frame loss can severely degrade semantic understanding, while adversarial attacks such as FGSM~\citep{goodfellow2014explaining}, PGD~\citep{madry2017towards}, and C\&W~\citep{carlini2017towards} can cause significant performance degradation even with small input perturbations. 
Benchmarks such as RoboBEV~\citep{xie2023robobev} have systematically exposed this fragility.

Existing robustness strategies provide partial solutions but still have limitations. 
Multi-modal fusion improves reliability by incorporating additional sensors, but it depends on extra sensor modalities and their reliability. 
Temporal aggregation methods, such as BEVFormer~\citep{li2022bevformer}, exploit historical information but may still propagate corrupted observations under persistent disturbances. 
Recent BEV refinement methods, such as DiffBEV~\citep{zou2024diffbev} and BEVDiffuser~\citep{ye2025bevdiffuser}, improve robustness by denoising corrupted BEV features, but diffusion-based refinement usually requires iterative sampling and introduces additional inference cost. 
These limitations motivate a lightweight and generalizable recovery mechanism that can exploit temporal structure without expensive iterative denoising.

\begin{figure}[t]
    \centering

    \begin{subfigure}[b]{0.58\linewidth}
        \centering
        \includegraphics[width=\linewidth]{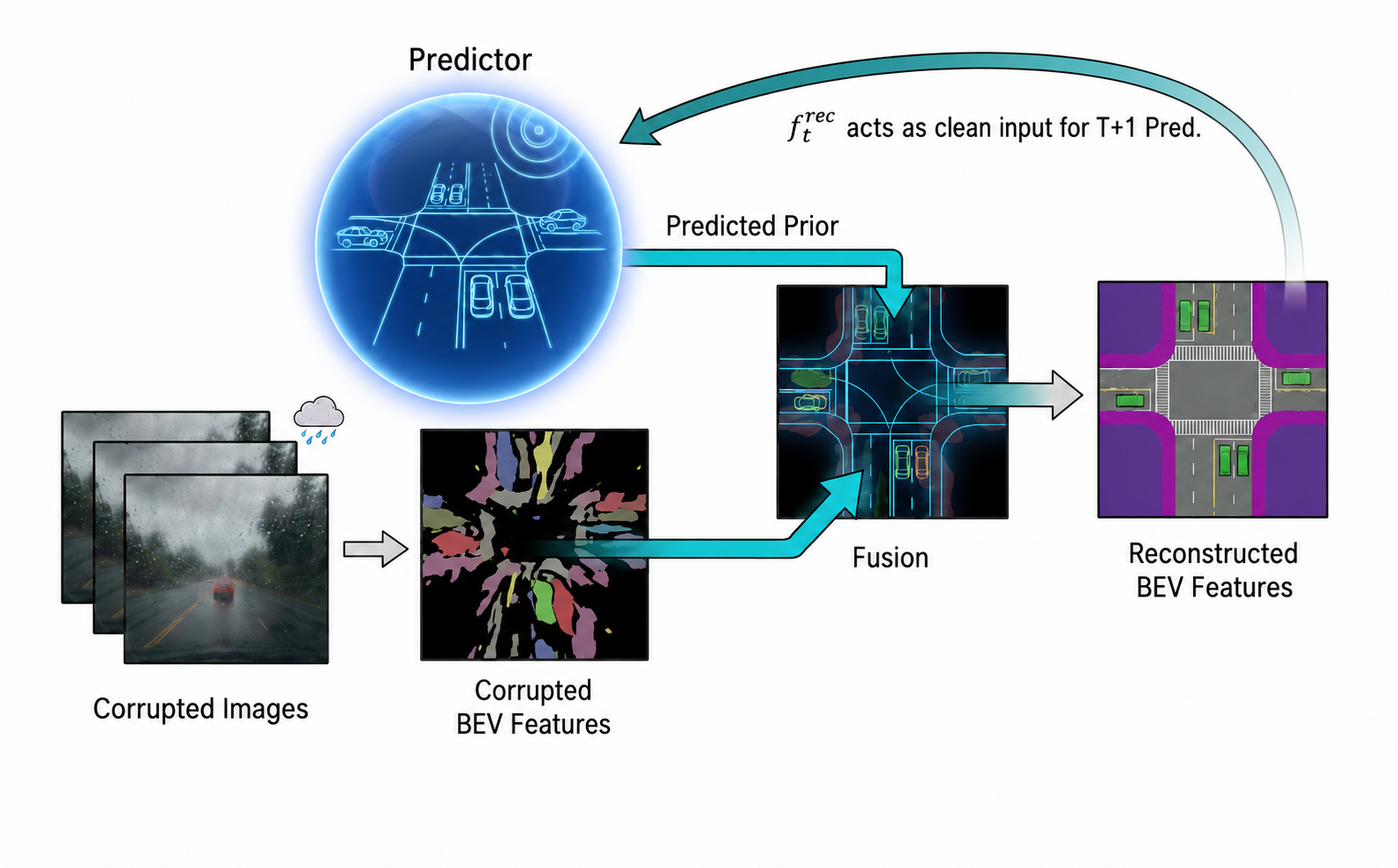}
        \caption{RESBev framework.}
        \label{fig:teaser_framework}
    \end{subfigure}
    \hfill
    \begin{subfigure}[b]{0.32\linewidth}
        \centering
        \includegraphics[width=\linewidth]{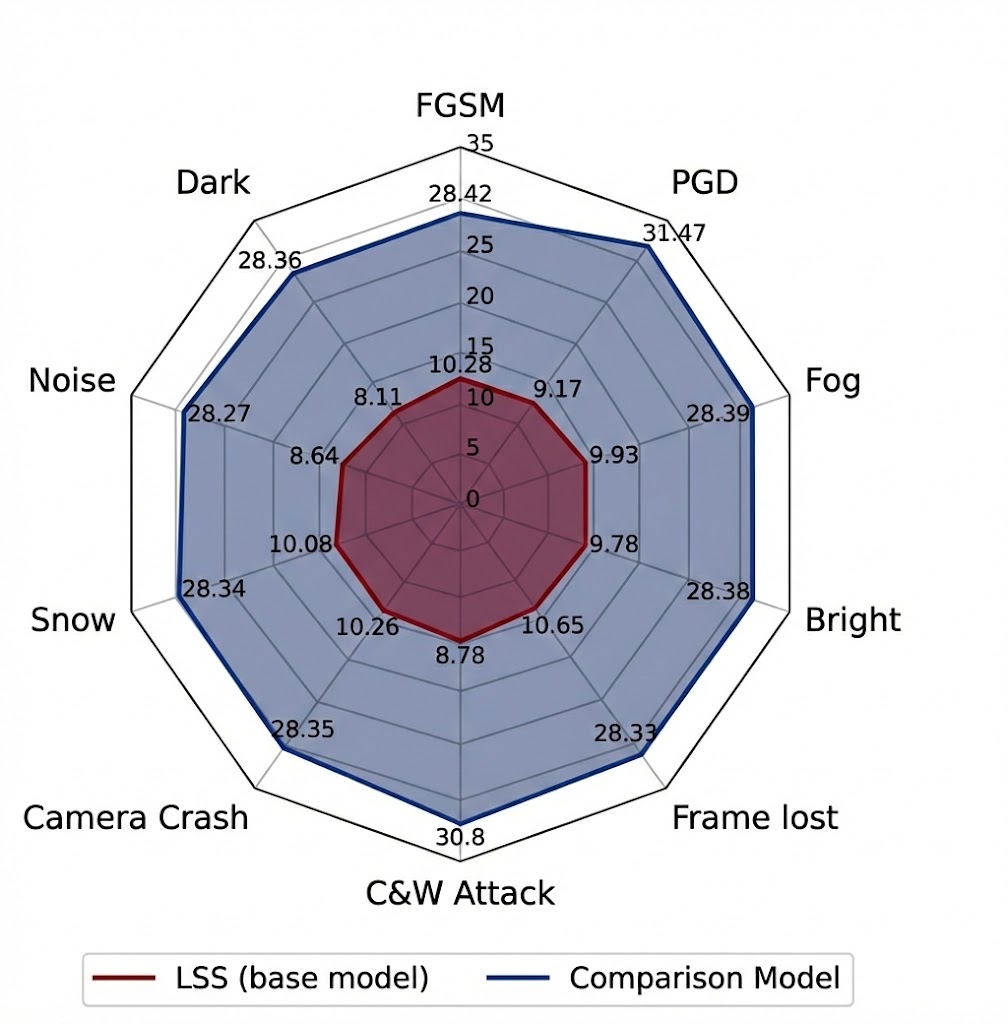}
        \caption{Robustness comparison.}
        \label{fig:teaser_radar}
    \end{subfigure}

    \caption{Overview and robustness comparison of RESBev. 
    \textbf{(a)} RESBev predicts a BEV prior from historical BEV features and ego-motion, and fuses it with corrupted observations to reconstruct the BEV representation.
    \textbf{(b)} RESBev improves the resilience of the LSS baseline under natural disturbances and adversarial attacks.}
    \label{fig:teaser}
\end{figure}

 Based on this insight, we introduce RESBev which incorporates latent dynamic predictor to learn the transition of BEV states over time (see Figure~\ref{fig:teaser_framework}). By modeling how BEV states evolve over time, RESBev can reconstruct corrupted features through temporal prior prediction rather than direct temporal aggregation. Extensive experiments demonstrate that RESBev can enhance the robustness of existing BEV models when facing various corruptions (see Figure~\ref{fig:teaser_radar}). Our contributions are summarized below:

\begin{enumerate}
\item We conduct a systematic analysis of robustness in BEV perception pipelines, revealing that effective recovery requires modeling in the BEV semantic space, preserving high-dimensional features before task compression, and leveraging temporal prior prediction.
\item Based on these insights, we propose a plug-and-play model that introduces a latent dynamic predictor to capture temporal dynamics in BEV semantic space and generate predictive semantic prior in complex traffic environments.
\item Extensive experiments demonstrate that the proposed method can effectively enhance the robustness of existing BEV perception methods against various disturbances and improve their generalization to unseen types of perception anomalies.
\end{enumerate}

\section{Related Work}

\subsection{BEV Perception}

In autonomous driving, bird's-eye-view (BEV) representation provides a unified environmental understanding necessary for downstream tasks. Lift-Splat-Shoot (LSS)~\citep{philion2020lift} establishes BEV representation by projecting multi-view 2D image features into a common BEV grid through explicit depth modeling. Building upon this paradigm, numerous BEV perception models have been proposed. For instance, GaussianLSS~\citep{sonarghare2025fisheyegaussianlift} refines depth estimation by incorporating Gaussian distributions to model uncertainty, while BEVFusion~\citep{liu2022bevfusion} integrates multi-modal sensor data to produce richer BEV representations. FIERY~\citep{hu2021fiery} extends the BEV framework to support multiple downstream tasks and leverages multi-frame temporal information within a unified architecture. Another line of work adopts Transformer-based architectures. For example, BEVFormer~\citep{li2022bevformer} introduces spatiotemporal transformers to aggregate features from historical frames and multiple camera views. Despite achieving strong performance on benchmarks such as nuScenes~\citep{caesar2020nuscenes}, these methods remain vulnerable to various real-world anomalies.

\subsection{Perception Anomalies}
Deep-learning-based perception systems for autonomous driving are highly vulnerable to various anomalies, including adversarial attacks and natural corruptions. 
Adversarial attacks involve crafting imperceptible perturbations to input data that induce incorrect predictions.
While classic digital attacks such as FGSM~\citep{goodfellow2014explaining}, PGD~\citep{madry2017towards}, and C\&W~\citep{carlini2017towards} were initially developed for 2D classification, recent studies demonstrate their degradation effect on BEV models \citep{zhu2023understanding, xie2023adversarial}. Beyond these, recent works have introduced targeted attacks for BEV perception, such as semantic perturbations through object swapping \citep{wang2025black} and multi-view adversarial attacks like ``Fool the Hydra''\citep{tarchoun2023fool}, which exploits vulnerabilities in cross-camera feature alignment. In addition, 3D adversarial examples using techniques like NeRF (e.g., Adv3D \citep{li2024adv3d}) create physically plausible perturbations. These attacks lead to perception failures such as missed detections or false positives. Beyond targeted attacks, real-world scenarios present diverse natural corruptions, including adverse weather (e.g., fog, dark, snow) and sensor failures (e.g., camera crash, frame loss).

\subsection{Robust Perception}

Ensuring robustness against real-world corruptions, such as adverse weather, sensor anomalies, and adversarial perturbations, is critical for autonomous driving perception. 
Recent studies~\citep{zhu2023understanding, xie2023adversarial} have shown that BEV perception pipelines are vulnerable to such corruptions, especially in modules such as view transformers. 
Benchmarks such as RoboBEV~\citep{xie2023robobev, xie2025benchmarking} further standardize robustness evaluation under diverse natural and sensor corruptions.

Existing defenses mainly fall into three categories. 
Multi-sensor fusion methods exploit complementary sensor information to improve reliability under adverse conditions~\citep{sadeghian2025reliability, kumar2025minimizing}. 
Temporal methods aggregate historical features to stabilize BEV representations under transient corruptions~\citep{li2022bevformer}. 
Diffusion-based BEV refinement methods have recently been used to improve BEV representation quality. 
DiffBEV~\citep{zou2024diffbev} adopts conditional diffusion to refine BEV features for BEV segmentation and 3D detection, while BEVDiffuser~\citep{ye2025bevdiffuser} uses ground-truth-layout-guided diffusion as a training-time plug-and-play denoising module for existing BEV detectors. 
These methods show the benefit of explicit BEV feature denoising, but they mainly learn denoising-based feature refinement and do not explicitly model temporal BEV state transitions under consecutive corruptions or adversarial disturbances.

In contrast, RESBev formulates robust BEV perception as \textit{temporal semantic prior prediction} followed by \textit{prior-guided reconstruction}. 
Unlike diffusion-based refinement methods such as DiffBEV~\citep{zou2024diffbev} and BEVDiffuser~\citep{ye2025bevdiffuser}, which primarily denoise single-frame BEV features through iterative or conditional diffusion, RESBev leverages historical reconstructed BEV states along with ego-motion to predict a clean BEV prior for the current timestamp. 
This predicted prior is then used to selectively retrieve reliable information from the corrupted observation, preserving temporal consistency across consecutive frames. 
Consequently, RESBev achieves robust feature recovery in a \textbf{single-step, non-iterative} manner, and can be seamlessly attached to dense BEV backbones such as LSS, SimpleBEV, GaussianLSS, and BEVFormer, without additional diffusion sampling overhead. 

\section{Analysis: Design Principles for Robust BEV Recovery}
\label{sec:analysis}

Robust BEV perception often degrades under real-world anomalies such as adverse weather, sensor corruption, and adversarial perturbations. 
Before introducing our method, we conduct diagnostic experiments to answer two questions: 
\emph{where} corrupted representations should be recovered, and \emph{how} temporal information should be used for recovery. 
This section is not intended to compare final systems, but to derive the design principles behind RESBev.

\paragraph{Diagnostic protocol.}
We use the LSS~\citep{philion2020lift} as a diagnostic testbed because it explicitly separates perspective-view feature extraction, BEV feature construction, and task prediction. This does not restrict our method to LSS-based models; the conclusion applies to any architecture with dense BEV features, such as BEVFormer.

\begin{figure}[tb]
    \centering
    \begin{subfigure}[b]{0.48\linewidth}
        \centering
        \includegraphics[width=\linewidth,height=0.9\linewidth,keepaspectratio]{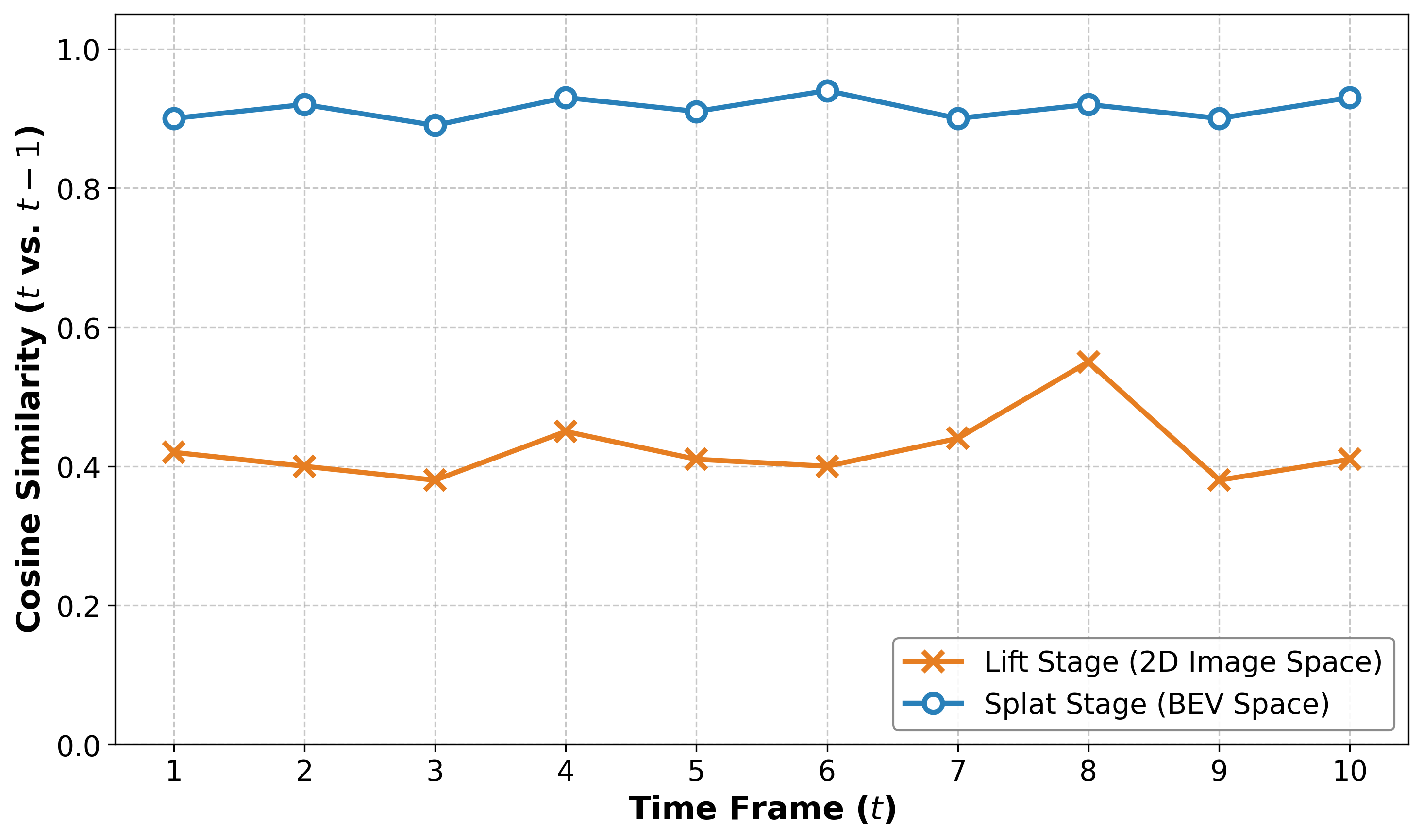}
        \caption{\textbf{Cross-frame stability.} BEV features remain more stable under persistent noise corruptions.}
        \label{fig:analysis_consistency}
    \end{subfigure}
    \hfill
    \begin{subfigure}[b]{0.48\linewidth}
        \centering
        \includegraphics[width=\linewidth,height=0.9\linewidth,keepaspectratio]{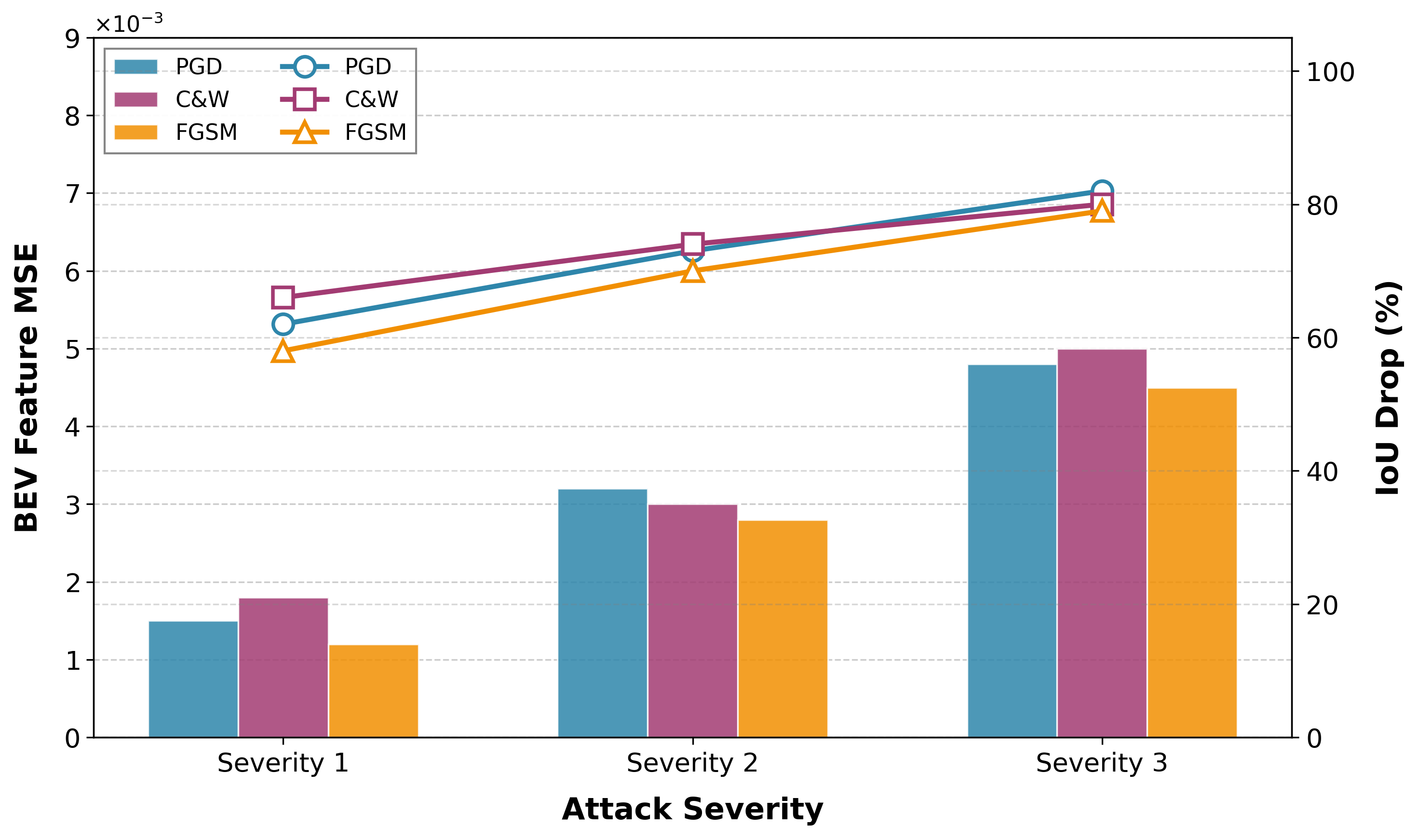}
        \caption{\textbf{Feature brittleness.} Small feature perturbations can cause large task drops.}
        \label{fig:analysis_perturbation}
    \end{subfigure}
    \caption{
    Diagnostic analysis for robust BEV recovery. 
    \textbf{(a)} BEV features show stronger temporal stability than perspective-view features. 
    \textbf{(b)} Small feature deviations can still cause severe performance degradation, motivating predictive recovery over simple aggregation.
    }
    \label{fig:analysis_combined}
\end{figure}

\subsection{Where should corrupted representations be recovered?}

The LSS pipeline contains three representative stages: perspective-view features after image encoding, BEV semantic features after view transformation, and task outputs after the prediction head. 
We compare recovery at the Lift, Splat, and Shoot stages under the same corruption setting.

\begin{table}[tb]
    \centering
    \caption{
    Recovery-stage diagnosis. 
    BEV features provide the best recovery performance under noise corruption.
    }
    \label{tab:stage_integration}
    \setlength{\tabcolsep}{4pt}
    \begin{tabular}{lccc}
        \toprule
        \textbf{Stage} & \textbf{Rep.} & \textbf{Feature Form} & \textbf{IoU} \\
        \midrule
        Lift & Perspective & $C \times H_{img} \times W_{img}$ & 16.4 \\
        Shoot & Task output & $N_{cls} \times X \times Y$ & 18.7 \\
        \textbf{Splat} & \textbf{BEV feature} & $C \times X \times Y$ & \textbf{31.6} \\
        \bottomrule
    \end{tabular}
\end{table}

Figure~\ref{fig:analysis_consistency} and Table~\ref{tab:stage_integration} show that BEV semantic features provide the most suitable recovery space. 
Perspective-view features are sensitive to image noise, viewpoint changes, and ego-motion, while task outputs are too compressed to retain the information needed for reconstruction. 
In contrast, Splat-stage BEV features remain high-dimensional, spatially structured, and temporally aligned, making them better suited for robust recovery.

\subsection{Is temporal aggregation sufficient?}

We next examine whether temporal aggregation alone can recover corrupted BEV features. 
All variants are implemented on the same Splat-stage BEV features using the same segmentation baseline and corruption setting as Table~\ref{tab:stage_integration}. 
The temporal attention variant follows the BEVFormer-style temporal attention design~\citep{li2022bevformer}, while the recurrent fusion variant follows the recurrent temporal fusion principle of RecurrentBEV~\citep{chang2024recurrentbev}. 
They are used as diagnostic variants rather than full system comparisons.

Table~\ref{tab:temporal_diagnostic} shows that temporal context helps, but direct aggregation remains limited because it still incorporates corrupted current features and may propagate anomalies. 
This is particularly critical under adversarial perturbations, where minor feature deviations may lead to substantial performance loss, as shown in Figure~\ref{fig:analysis_perturbation}. 
In contrast, the predictive-prior variant estimates the current BEV state from historical BEV features and ego-motion, providing a cleaner reference for robust reconstruction.

\begin{table}[h]
    \centering
    \caption{
    Temporal-mechanism diagnosis on the same BEV features. 
    Predicting a temporal prior is more effective than directly aggregating corrupted observations.
    }
    \label{tab:temporal_diagnostic}
    \setlength{\tabcolsep}{4pt}
    \begin{tabular}{lcccc}
        \toprule
        \textbf{Variant} & \textbf{Hist.} & \textbf{Curr.} & \textbf{Prior} & \textbf{IoU} \\
        \midrule
        Single frame & No & Yes & No & 15.53 \\
        Temporal attn. & Yes & Yes & No & 20.17 \\
        Recurrent fusion & Yes & Yes & No & 22.31 \\
        Predictive prior & Yes & No & Yes & \textbf{30.11} \\
        \bottomrule
    \end{tabular}
\end{table}

\paragraph{Design implications.}
These findings lead to two design requirements. 
First, recovery should be performed in dense BEV semantic feature space. 
Second, historical information should be used to predict a temporal prior rather than being directly aggregated with corrupted current observations. 
These requirements motivate RESBev, which implements prior prediction and prior-guided reconstruction in the BEV feature space.

\begin{figure*}[h]
    \centering
    \includegraphics[width=0.8\textwidth]{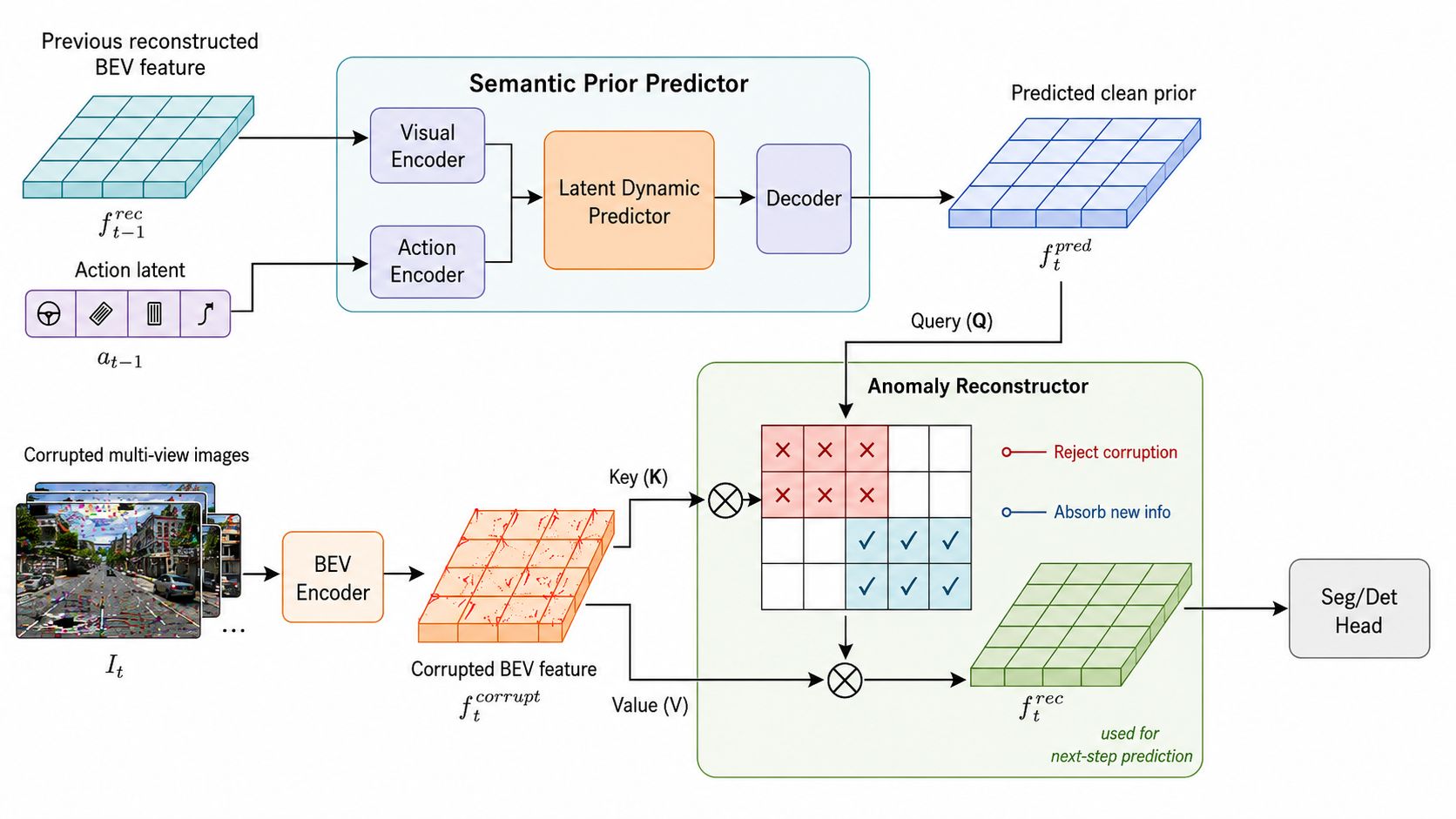}
    \caption{The overall architecture of our proposed temporal fusion model. The model consists of two core components: Semantic Prior Predictor that predicts the current BEV state from the past, and Anomaly Reconstructor that fuses this prediction with current fused BEV features.}
    
    \label{fig:overall_architecture}
\end{figure*}

\section{Method}
\label{sec:method}
\subsection{Overall Framework}

RESBev consists of two modules, as shown in Figure~\ref{fig:overall_architecture}: Semantic Prior Predictor and Anomaly Reconstructor. 
Given the previous reconstructed BEV feature and ego-motion, the Semantic Prior Predictor predicts a BEV prior for current timestamp. 
The Anomaly Reconstructor then uses this predicted prior as a query to extract reliable information from current corrupted BEV feature through cross-attention. 
The final reconstructed feature is obtained by combining the predicted prior with the retrieved current-frame information through a learnable gate. 
This feature is then fed into the downstream task head.

\subsubsection{Semantic Prior Predictor.}

This module is trained to predict the future state. Given the previous reconstructed feature ($f^{rec}_{t-1}$) and the ego-vehicle's motion, a multi-modal encoder is used to extract semantic BEV latents and ego-motion latents for short-term temporal prediction. Subsequently, a latent dynamic predictor is introduced to predict the current BEV prior from historical BEV features and ego-motion, which predicts the future BEV prior ($f^{pred}_{t}$) for the anomaly recovery and robust perception.

\subsubsection{Anomaly Reconstructor.}

This module is trained to recover the final BEV features $f^{rec}_{t}$ by fusing the predicted prior $f^{pred}_{t}$ with the current corrupted BEV feature $f^{corrupt}_{t}$. 
Using $f^{pred}_{t}$ as a semantic reference, it selectively extracts reliable information that is consistent with the predicted scene prior, while suppressing anomalous responses caused by corruptions. 
This produces a reconstructed BEV representation that preserves temporal consistency while remaining responsive to valid scene changes.

\subsection{Semantic Prior Predictor}
\label{sec:latent_dynamics_module}
To restore corrupted observations, we introduce the Semantic Prior Predictor. As illustrated in our pipeline, this module leverages historical information to predict BEV representations, serving as a prior for fusion. Instead of computing transitions directly in the high-dimensional dense feature space, we project the inputs into a compact latent space. Specifically, given the reconstructed BEV features $f^{rec}_{t-1}$ from the previous step and corresponding ego-vehicle motion vector $a_{t-1}$ (comprising translation and rotation), we employ a visual encoder $E_{vis}$ and an action encoder $E_{act}$ to extract visual and action latents. These latents are concatenated to form an action-aware visual latent state. A Transformer-based latent dynamic predictor (LDP) then models the spatiotemporal transition to predict the future latent state at time $t$. Finally, a decoder $D$ maps this predicted latent back to the dense BEV feature space to obtain the predicted prior $f^{pred}_{t}$. The entire generation process of this prior can be formulated as:
\begin{equation}
f^{pred}_{t} = D\Big(\text{LDP}\big(\text{Concat}(E_{vis}(f^{rec}_{t-1}), E_{act}(a_{t-1}))\big)\Big).
\end{equation}

\subsection{Anomaly Reconstructor}\label{sec:temporal_cross_attention}

While the latent dynamic predictor provides a historical prior, it cannot predict sudden stochastic events, such as a car unexpectedly driving into the scenario. To ground this prediction in real-time reality without absorbing current noise, we introduce a temporal cross attention module. Specifically, given the current multi-view images at time $t$, which may suffer from various anomalies, we process them through the BEV encoder to obtain the current corrupted BEV representation, denoted as $f^{corrupt}_{t}$. 

To enhance temporal stability and provide a reliable reference for reconstruction, we formulate this process as a query-driven extraction that uses the predicted prior to retrieve valid information from the current corrupted observation. 
Specifically, the predicted feature $f^{pred}_{t}$ acts as the query ($Q$), while the current corrupted BEV feature $f^{corrupt}_{t}$ serves as both the Key ($K$) and Value ($V$). 
This design allows the model to selectively extract perceptual details that are consistent with the predicted prior, while suppressing anomalous responses in the corrupted observation. 
To integrate the retrieved information without destabilizing the prior, we employ a dynamically gated residual connection:

\begin{equation}\label{eq:fusion_v2}
f^{rec}_{t} = f^{pred}_{t} + \alpha \cdot \mathrm{CrossAttn}\left(f^{pred}_{t}, f^{corrupt}_{t}\right),
\end{equation}
where $\alpha \in [0, 1]$ is a learned dynamic gating factor. This gate adaptively controls the information flow, forcing the model to rely more on the historical prior when the current observation is heavily corrupted, while attending to $f^{corrupt}_{t}$ when it introduces valuable new context.

\subsection{Training Objective}
\label{sec:training_objective}

We train RESBev with two-frame multi-view inputs from the same driving scene. 
At each timestamp, the input contains six camera views, denoted as $\mathbf{I}_t=\{I_t^v\}_{v=1}^{6}$. 
For each pair $(\mathbf{I}_{t-1}, \mathbf{I}_t)$, the current input is always corrupted, while the historical input is randomly kept clean or corrupted. 
This exposes the model to both clean-start and corrupted-start conditions, while the clean current feature is used only as supervision.

The corrupted current input is:
\begin{equation}
\mathbf{I}^{corrupt}_{t}=\mathcal{C}_{t}(\mathbf{I}_{t}),
\end{equation}
where $\mathcal{C}_{t}(\cdot)$ denotes a randomly sampled corruption. 
The historical input is sampled as:
\begin{equation}
\mathbf{I}^{hist}_{t-1}
=
\begin{cases}
\mathbf{I}_{t-1}, & m=1,\\
\mathcal{C}_{t-1}(\mathbf{I}_{t-1}), & m=0,
\end{cases}
\qquad
m \sim \mathrm{Bernoulli}(p_{hist}),
\end{equation}
where $p_{hist}$ controls the probability of using a clean historical input.

The BEV encoder extracts:
\begin{equation}
f^{hist}_{t-1}=B(\mathbf{I}^{hist}_{t-1}),\quad
f^{corrupt}_{t}=B(\mathbf{I}^{corrupt}_{t}),\quad
f^{clean}_{t}=B(\mathbf{I}_{t}),
\end{equation}
where $B(\cdot)$ denotes the backbone and view transformation module.

Given $f^{hist}_{t-1}$ and ego-motion $a_{t-1}$, the Latent Dynamic Predictor estimates the current BEV prior, the Anomaly Reconstructor recovers the current BEV feature, and the clean BEV feature is used as the stop-gradient target:
\begin{equation}
f^{pred}_{t}=\mathcal{P}_{\theta}(f^{hist}_{t-1},a_{t-1}),\quad
f^{rec}_{t}=\mathcal{R}_{\phi}(f^{pred}_{t},f^{corrupt}_{t}),\quad
f^{target}_{t}=\mathrm{sg}(f^{clean}_{t}).
\end{equation}

We optimize the prior prediction loss, reconstruction loss, and task loss:
\begin{equation}
\mathcal{L}_{p}=\mathrm{MSE}(f^{pred}_{t},f^{target}_{t}),\quad
\mathcal{L}_{r}=\mathrm{MSE}(f^{rec}_{t},f^{target}_{t}),\quad
\mathcal{L}_{task}=\ell_{task}(H(f^{rec}_{t}),y_t).
\end{equation}
Here, $\mathcal{L}_{p}$, $\mathcal{L}_{r}$, and $\mathcal{L}_{task}$ denote the prior prediction, feature reconstruction, and downstream task losses, respectively.

The final objective is:
\begin{equation}
\mathcal{L}
=
\mathcal{L}_{task}
+
\lambda_{p}\mathcal{L}_{p}
+
\lambda_{r}\mathcal{L}_{r}.
\end{equation}

During inference, only one forward pass is required. 
For single-step inference, the LDP uses the historical BEV feature extracted from the previous multi-view input. 
For recursive long-horizon inference, the previous reconstructed feature $f^{rec}_{t-1}$ is used as the historical BEV state for predicting the next prior.

\section{Experiments}
This section validates the effectiveness of RESBev in improving robust BEV perception. We first measure performance on benchmark corruptions, then assess generalization to unseen corruptions. We then provide component ablations, long-horizon robustness analysis, and deployment cost measurements.

\subsection{Experimental Setup}

\subsubsection{Dataset and Baseline Models.} 
Our experiments are conducted on the widely used nuScenes dataset~\citep{caesar2020nuscenes}. We evaluate our method on the task of bird's-eye-view (BEV) semantic segmentation and 3D object detection. To validate the performance of our RESBev model, we integrate it into three LSS-based models: LSS~\citep{philion2020lift}, SimpleBEV~\citep{harley2022simple}, GaussianLSS~\citep{sonarghare2025fisheyegaussianlift}, and one attention-based model: BEVFormer~\citep{li2022bevformer}. To further benchmark our method against state-of-the-art robust perception frameworks, we also compare it with BEVDiffuser~\citep{ye2025bevdiffuser} and DiffBEV~\citep{zou2024diffbev}.

\subsubsection{Robustness Evaluation Protocol.}
We adopt the evaluation protocol from RoboBEV~\citep{xie2023robobev}, testing our models against ten types of corruptions at three severity levels. The severity levels are carefully determined to avoid excessive performance drops that could undermine the reliability of our conclusions. Due to space limitations, detailed specifications for each corruption severity level are provided in the supplementary material. These corruptions are categorized as follows:
\begin{itemize}
    \item Natural Corruptions: We simulate diverse environmental conditions and sensor failures, including fog, snow, bright, dark, camera crash, frame loss, and noise.
    \item Adversarial Attacks: To measure resilience against malicious inputs, we employ three attacks designed for BEV models: Fast Gradient Sign Method (FGSM)~\citep{goodfellow2014explaining}, Projected Gradient Descent (PGD)~\citep{madry2017towards}, and Carlini \& Wagner (C\&W) ~\citep{carlini2017towards}.
\end{itemize}

\subsubsection{Implementation Details and Metrics.} 
We report IoU for the vehicle class in BEV semantic segmentation and NDS (nuScenes Detection Score) for 3D object detection.  Our method is implemented using PyTorch. All models are trained on a NVIDIA A100-SXM4-80GB GPU with a batch size of 16.

\begin{table*}[tb]
\centering
\renewcommand{\arraystretch}{1.2} 
\caption{
Average IoU under seen corruptions across three severity levels. 
The Change column denotes the absolute IoU improvement of RESBev over the vanilla model.
}
\label{tab:main_results_table}
\resizebox{\textwidth}{!}{%
\begin{tabular}{l 
c c >{\columncolor{gray!15}}c 
c c >{\columncolor{gray!15}}c 
c c >{\columncolor{gray!15}}c 
c c >{\columncolor{gray!15}}c 
c
>{\columncolor{gray!15}}c}
\toprule
\textbf{Corruption} 
& \multicolumn{3}{c}{\textbf{LSS~\citep{philion2020lift}}}
& \multicolumn{3}{c}{\textbf{SimpleBEV~\citep{harley2022simple}}}
& \multicolumn{3}{c}{\textbf{GaussianLSS~\citep{sonarghare2025fisheyegaussianlift}}}
& \multicolumn{3}{c}{\textbf{BEVFormer~\citep{li2022bevformer}}}
&
& \multicolumn{1}{c}{\textbf{DiffBEV~\citep{zou2024diffbev}}} \\
\cmidrule(lr){2-4}
\cmidrule(lr){5-7}
\cmidrule(lr){8-10}
\cmidrule(lr){11-13}
\cmidrule(lr){15-15}

& Vanilla & +RESBev & \textbf{Change}
& Vanilla & +RESBev & \textbf{Change}
& Vanilla & +RESBev & \textbf{Change}
& Vanilla & +RESBev & \textbf{Change}
&
& Vanilla \\
\midrule

\textbf{Clean}
& 33.03 & 33.31 & $\uparrow$\textbf{0.28}
& 55.70 & 55.91 & $\uparrow$\textbf{0.21}
& 42.80 & 43.14 & $\uparrow$\textbf{0.34}
& 46.70 & 47.01 & $\uparrow$\textbf{0.31}
&
& 38.90 \\

\midrule
\textbf{FGSM}
& 10.28 & 28.42 & $\uparrow$\textbf{18.14}
& 13.23 & 31.09 & $\uparrow$\textbf{17.86}
& 9.91  & 31.43 & $\uparrow$\textbf{21.52}
& 21.54 & 38.78 & $\uparrow$\textbf{17.24}
&
& 25.18 \\

\textbf{PGD}
& 9.17 & 31.47 & $\uparrow$\textbf{22.30}
& 11.89 & 31.65 & $\uparrow$\textbf{19.76}
& 7.72 & 31.09 & $\uparrow$\textbf{23.37}
& 18.68  & 38.90 & $\uparrow$\textbf{20.22}
&
& 28.11 \\

\textbf{Fog}
& 9.93 & 28.39 & $\uparrow$\textbf{18.46}
& 17.18 & 33.76 & $\uparrow$\textbf{16.58}
& 24.26 & 31.95 & $\uparrow$\textbf{7.69}
& 25.34 & 38.91 &$\uparrow$\textbf{13.57}
&
& 29.02 \\

\textbf{Bright}
& 9.78 & 28.38 & $\uparrow$\textbf{18.60}
& 17.63 & 33.79 & $\uparrow$\textbf{16.16}
& 24.03 & 31.87 & $\uparrow$\textbf{7.85}
& 25.70 & 38.74 & $\uparrow$\textbf{13.04}
&
& 30.39 \\

\textbf{Frame Lost}
& 10.65 & 28.33 & $\uparrow$\textbf{17.68}
& 17.74 & 33.73 & $\uparrow$\textbf{15.99}
& 14.70 & 31.61 & $\uparrow$\textbf{16.90}
& 25.86 & 38.75 & $\uparrow$\textbf{12.89}
&
& 27.81 \\

\midrule
\textbf{Overall Avg.}
& 9.96 & 29.02 & \textbf{$\uparrow$19.06}
& 15.53 & 32.80 & \textbf{$\uparrow$17.27}
& 16.12 & 31.59 & \textbf{$\uparrow$15.47}
& 23.42 & 38.82 & \textbf{$\uparrow$15.40}
&
& 28.10 \\
\bottomrule
\end{tabular}%
}
\end{table*}

\begin{table*}[tb]
    \centering
    \caption{
    Average NDS under seen and unseen corruptions across three severity levels. 
    BEVDiffuser and RESBev are evaluated on the same BEVFormer and BEVFusion backbones.
    }
    \label{tab:nds_seen_unseen}

    \scriptsize
    \setlength{\tabcolsep}{1.6pt}
    \renewcommand{\arraystretch}{0.82}

    \begin{subtable}{0.49\textwidth}
        \centering
        \caption{Seen corruptions.}
        \label{tab:nds_seen}
        \scalebox{0.74}{
        \begin{tabular}{lcccccc}
            \toprule
            \textbf{Corruption}
            & \multicolumn{3}{c}{\textbf{BEVFormer-base}}
            & \multicolumn{3}{c}{\textbf{BEVFusion}} \\
            \cmidrule(lr){2-4}
            \cmidrule(lr){5-7}
            & Vanilla & +RESBev & +BEVDiffuser
            & Vanilla & +RESBev & +BEVDiffuser \\
           \midrule
            \textbf{Clean} & 51.8 & 53.3 & 53.7 & 70.9 & 71.7 & 71.9 \\
            \midrule
            FGSM       & 30.10 & 50.21 & 44.31 & 52.18 & 70.81 & 66.71 \\
            PGD        & 28.18 & 50.19 & 40.87 & 51.09 & 70.73 & 64.90 \\
            Fog        & 34.81 & 51.35 & 45.71 & 52.83 & 71.98 & 68.71 \\
            Bright     & 40.31 & 51.98 & 44.84 & 52.19 & 70.19 & 69.15 \\
            Frame Lost & 35.79 & 51.81 & 44.69 & 52.87 & 70.54 & 64.83 \\
            \midrule
            \textbf{Avg.} & \textbf{33.84} & \textbf{51.11} & \textbf{44.08} & \textbf{52.23} & \textbf{70.85} & \textbf{66.86} \\
            \bottomrule
        \end{tabular}
        }
    \end{subtable}
    \hfill
    \begin{subtable}{0.49\textwidth}
        \centering
        \caption{Unseen corruptions.}
        \label{tab:nds_unseen}
        \scalebox{0.74}{
        \begin{tabular}{lcccccc}
            \toprule
            \textbf{Corruption}
            & \multicolumn{3}{c}{\textbf{BEVFormer-base}}
            & \multicolumn{3}{c}{\textbf{BEVFusion}} \\
            \cmidrule(lr){2-4}
            \cmidrule(lr){5-7}
            & Vanilla & +RESBev & +BEVDiffuser
            & Vanilla & +RESBev & +BEVDiffuser \\
            \midrule
            \textbf{Clean} & 51.8 & 53.3 & 53.7 & 70.9 & 71.7 & 71.9 \\
            \midrule
            C\&W       & 22.81 & 49.88 & 38.18 & 45.18 & 70.84 & 60.91 \\
            Snow       & 30.19 & 50.92 & 43.71 & 51.91 & 71.92 & 64.87 \\
            Cam. Crash & 29.84 & 51.15 & 44.87 & 52.35 & 70.63 & 62.95 \\
            Noise      & 29.97 & 50.87 & 45.14 & 53.14 & 70.95 & 65.73 \\
            Dark       & 26.71 & 51.32 & 44.91 & 52.84 & 70.28 & 64.80 \\
            \midrule
            \textbf{Avg.} & \textbf{27.90} & \textbf{50.83} & \textbf{43.36} & \textbf{51.08} & \textbf{70.92} & \textbf{63.85} \\
            \bottomrule
        \end{tabular}
        }
    \end{subtable}
\end{table*}

\subsection{Performance on Benchmark Corruptions}

We evaluate RESBev on five benchmark corruptions and report average IoU and average NDS across three severity levels. 
Detailed per-severity results are provided in the supplementary material. 
As shown in Table~\ref{tab:main_results_table} and~\ref{tab:nds_seen}, vanilla BEV models degrade severely under corruptions, while adding RESBev consistently improves robustness across all evaluated backbones. 
Compared with DiffBEV, RESBev maintains stronger corrupted-scene performance with a lightweight prior-guided reconstruction design.

\subsection{Generalization to Unseen Corruptions}

We evaluate whether RESBev generalizes to corruptions not seen during training. 
Models are trained on five corruptions and tested on five held-out corruptions. 
As shown in Table~\ref{tab:additional_corruptions} and~\ref{tab:nds_unseen}, vanilla models suffer large performance drops, while RESBev consistently improves all backbones. 
The strong unseen-corruption results indicate that RESBev learns temporal BEV structure rather than overfitting to specific corruption types.

\begin{table*}[tb]
    \centering
    \renewcommand{\arraystretch}{1.2} 
    \caption{
    Average IoU under unseen corruptions across three severity levels. 
    }
    \label{tab:additional_corruptions}
    \resizebox{\textwidth}{!}{%
    \begin{tabular}{l 
    c c >{\columncolor{gray!15}}c 
    c c >{\columncolor{gray!15}}c 
    c c >{\columncolor{gray!15}}c 
    c c >{\columncolor{gray!15}}c 
    c 
    >{\columncolor{gray!15}}c}
        \toprule
        \textbf{Corruption} 
        & \multicolumn{3}{c}{\textbf{LSS~\citep{philion2020lift}}} 
        & \multicolumn{3}{c}{\textbf{SimpleBEV~\citep{harley2022simple}}} 
        & \multicolumn{3}{c}{\textbf{GaussianLSS~\citep{sonarghare2025fisheyegaussianlift}}} 
        & \multicolumn{3}{c}{\textbf{BEVFormer~\citep{li2022bevformer}}} 
        &
        & \multicolumn{1}{c}{\textbf{DiffBEV~\citep{zou2024diffbev}}} \\
        
        \cmidrule(lr){2-4} 
        \cmidrule(lr){5-7} 
        \cmidrule(lr){8-10} 
        \cmidrule(lr){11-13} 
        \cmidrule(lr){15-15}
        
        & Vanilla & +RESBev & \textbf{Change} 
        & Vanilla & +RESBev & \textbf{Change} 
        & Vanilla & +RESBev & \textbf{Change} 
        & Vanilla & +RESBev & \textbf{Change} 
        &
        & Vanilla \\
        \midrule
        
        \textbf{Clean}       
        & 33.03 & 33.31 & $\uparrow$\textbf{0.28}
        & 55.70 & 55.91 & $\uparrow$\textbf{0.21}
        & 42.80 & 43.14 & $\uparrow$\textbf{0.34}
        & 46.70 & 47.01 & $\uparrow$\textbf{0.31}
        & 
        & 38.90 \\
        \midrule
        
        \textbf{C\&W Attack} 
        & 8.78 & 30.80 & $\uparrow$\textbf{22.02} 
        & 11.43 & 30.61 & $\uparrow$\textbf{19.18} 
        & 5.97 & 31.24 & $\uparrow$\textbf{25.27} 
        & 17.56 & 39.67 & $\uparrow$\textbf{22.11}
        & 
        & 29.81 \\
        
        \textbf{Snow}        
        & 10.26 & 28.35 & $\uparrow$\textbf{18.09} 
        & 17.40 & 33.71 & $\uparrow$\textbf{16.31} 
        & 16.08 & 32.10 & $\uparrow$\textbf{16.02} 
        & 28.11 & 38.91 & $\uparrow$\textbf{10.80}
        & 
        & 30.14 \\
        
        \textbf{Camera Crash}
        & 10.08 & 28.34 & $\uparrow$\textbf{18.27} 
        & 17.34 & 33.74 & $\uparrow$\textbf{16.40} 
        & 13.41 & 31.56 & $\uparrow$\textbf{18.15} 
        & 27.46  & 39.72  & $\uparrow$\textbf{12.26}
        & 
        & 29.42 \\
        
        \textbf{Noise}       
        & 8.64 & 28.27 & $\uparrow$\textbf{19.63} 
        & 17.60 & 33.76 & $\uparrow$\textbf{16.16} 
        & 16.67 & 31.43 & $\uparrow$\textbf{14.76} 
        & 21.55 & 39.34  & $\uparrow$\textbf{17.79}
        & 
        & 29.89 \\
        
        \textbf{Dark}        
        & 8.11 & 28.36 & $\uparrow$\textbf{20.25} 
        & 16.29 & 33.75 & $\uparrow$\textbf{17.46} 
        & 17.68 & 31.96 & $\uparrow$\textbf{14.28} 
        & 28.06 & 39.75 & $\uparrow$\textbf{11.69}
        & 
        & 30.10 \\
        \midrule
        
        \textbf{Overall Avg.} 
        & 9.17 & 28.82 & \textbf{$\uparrow$19.65} 
        & 16.01 & 33.11 & \textbf{$\uparrow$17.10} 
        & 13.96 & 31.66 & \textbf{$\uparrow$17.70} 
        & 24.55 & 39.48 & \textbf{$\uparrow$14.93} 
        & 
        & 29.87 \\
        \bottomrule
    \end{tabular}%
    }
\end{table*}

\subsection{Ablation Study}

Table~\ref{tab:ablation_study_single_column} evaluates the contributions of the Latent Dynamic Predictor and Anomaly Reconstructor. 
Using the predictor alone provides a strong temporal prior, while adding the reconstructor consistently improves IoU across all baselines. 
This shows that prior-guided reconstruction is important for incorporating valid current-frame information while suppressing corruptions.

\begin{table}[t]
    \centering
    \caption{
    Ablation study of RESBev components. 
    SBEV denotes SimpleBEV and GLSS denotes GaussianLSS. 
    All results are mean IoU over five seen corruptions, where each sample is evaluated under a randomly selected corruption type and severity. 
    Gain is computed over the Predictor-only baseline.
    }
    \label{tab:ablation_study_single_column}
    \setlength{\tabcolsep}{4pt}
    \begin{tabular}{lcccc}
        \toprule
        \textbf{Config.} & \textbf{LSS} & \textbf{SBEV} & \textbf{GLSS} & \textbf{BEVFormer} \\
        \midrule
        Predictor-only & 26.67 & 30.11 & 29.16 & 35.48 \\
        Predictor + Reconstructor & 29.02 & 32.80 & 31.59 & 38.82 \\
        \midrule
        \rowcolor{gray!15} 
        \textbf{Gain} & \textbf{$\uparrow$8.8\%} & \textbf{$\uparrow$8.9\%} & \textbf{$\uparrow$8.3\%} & \textbf{$\uparrow$9.4\%} \\
        \bottomrule
    \end{tabular}
\end{table}

\subsection{Robustness to Consecutive Corruptions}

Real-world perception systems may face persistent disturbances rather than isolated corruptions. 
We therefore evaluate RESBev with an $n$-step recursive reconstruction task using LSS~\citep{philion2020lift}, where each reconstructed feature is reused as the historical input for the Latent Dynamic Predictor at the next step.
We use a 50-step setting ($n=50$), where all frames in one sequence are corrupted by the selected corruption.
As shown in Table~\ref{tab:consecutive_corruptions_comparison}, RESBev maintains stable IoU under consecutive seen corruptions, indicating limited error accumulation under continuous disturbances. 
Additional results on unseen corruptions are provided in the supplementary material.

\begin{table}[t]
    \centering
    \small
    \caption{
    Long-horizon robustness under consecutive seen corruptions.
    }
    \label{tab:consecutive_corruptions_comparison}
    \setlength{\tabcolsep}{3.5pt}
    \resizebox{0.7\textwidth}{!}{%
    \begin{tabular}{l c c c c >{\columncolor{gray!15}}c}
        \toprule
        \textbf{Corruption} & \textbf{1-Step} & \textbf{10-Step} & \textbf{30-Step} & \textbf{50-Step} & \textbf{Change} \\
        \midrule
        FGSM & 28.42 & 28.58 & 28.12 & 28.11 & $\downarrow$1.09\% \\
        PGD & 31.47 & 30.93 & 30.97 & 30.84 & $\downarrow$2.00\% \\
        Fog & 28.39 & 28.03 & 28.00 & 27.89 & $\downarrow$1.76\% \\
        Bright & 28.38 & 27.91 & 27.88 & 27.89 & $\downarrow$1.72\% \\
        Frame Lost & 28.33 & 28.48 & 28.31 & 28.43 & $\uparrow$0.35\% \\
        \bottomrule
    \end{tabular}%
    }
\end{table}

\subsection{Efficiency and Deployment Cost Analysis}
\label{sec:efficiency_analysis}

We compare RESBev with BEVFormer-base under the same input resolution and hardware setting. FPS is computed from the average per-frame latency measured with batch size 1 after 100 warm-up iterations and 500 timed iterations using \texttt{torch.cuda.synchronize()}. RESBev reduces FPS only from 2.7 Hz to 2.4 Hz, indicating limited deployment overhead.

\begin{table}[h]
    \centering
    \caption{
    Efficiency and GPU memory usage comparison under FP16 inference.
    }
    \label{tab:efficiency_comparison}
    \setlength{\tabcolsep}{5pt} 
    {\large 
    \resizebox{0.85\textwidth}{!}{%
    \begin{tabular}{l c c c c c}
        \toprule
        \textbf{Method} & \textbf{Params (M)} & \textbf{FLOPs (G)} & \textbf{FPS (Hz)} & \textbf{Training Memory (GB)} & \textbf{Inference Memory (GB)} \\
        \midrule
        BEVFormer-base & 69.1 & 1311 & 2.7 & 28.5 & 8.4 \\
        BEVFormer-base + RESBev & 83.8 & 1552 & 2.4 & 32.1 & 9.8 \\
        \bottomrule
    \end{tabular}%
    }
    }
\end{table}

\section{Conclusion}

We present RESBev, a plug-and-play framework for robust BEV perception under natural corruptions and adversarial attacks. 
Rather than directly aggregating corrupted observations, RESBev predicts a temporal BEV prior from historical context and ego-motion using a Latent Dynamic Predictor, and reconstructs the current BEV feature through a prior-guided Anomaly Reconstructor. 
Experiments on nuScenes show that RESBev improves robustness across multiple BEV models, generalizes to unseen corruptions, supports 3D detection, and maintains stability under long-horizon consecutive corruptions.

\bibliographystyle{plainnat}
\bibliography{references}


\appendix

\section{Technical appendices and supplementary material}

\noindent This supplementary material is organized as follows:\\
Sec. \ref{sec:supp_benchmark}: Benchmark and Corruption Details.\\
Sec. \ref{sec:supp_impl}: Implementation Details.\\
Sec. \ref{sec:supp_exp}: Additional Experiments.\\
Sec. \ref{sec:supp_discussion}: Discussions.\\
A supplementary video showing the real-time performance of RESBev is attached alongside this supplementary document.

\section{Benchmark and Corruption Details}
\label{sec:supp_benchmark}

\subsection{Benchmark Setup}
We follow the RoboBEV-style evaluation setting~\citep{xie2023robobev,xie2025benchmarking} and study robustness under both natural corruptions and adversarial attacks. During training, we use five corruptions: FGSM, PGD, Fog, Bright, and Frame Lost. These training-time corruptions are used to expose the model to representative disturbances of both adversarial and natural categories.

During evaluation, we report performance not only on the corruptions used during training, but also on five additional corruptions that are never introduced during training: C\&W Attack, Snow, Camera Crash, Noise, and Dark. This protocol allows us to evaluate both robustness to training-time corruptions and generalization to unseen test-time corruptions. Each corruption is evaluated under three severity levels.

\subsection{Implementation Details of Natural Corruptions}
We use seven natural corruptions in total: Noise, Bright, Dark, Fog, Snow, Camera Crash, and Frame Lost. The implementation details of each corruption are given below.

\paragraph{Noise.}
Noise is implemented as additive Gaussian perturbation. The standard deviation is defined relative to the dynamic range of the current image as $\sigma \in \{0.1 \times (\max(I)-\min(I)),\ 0.2 \times (\max(I)-\min(I)),\ 0.3 \times (\max(I)-\min(I))\}$ for severity levels 1, 2, and 3.

\paragraph{Bright.}
Bright corruption is implemented by directly rescaling the image intensity with a multiplicative brightness factor. The three severity levels use factors $\{1.3,\ 1.8,\ 2.2\}$.

\paragraph{Dark.}
Dark corruption is implemented by directly rescaling the image intensity with a multiplicative darkness factor. The three severity levels use factors $\{0.85,\ 0.325,\ 0.3\}$.

\paragraph{Fog.}
Fog is synthesized using a transmission-based atmospheric scattering model. The fog density parameter is set to $\{0.3,\ 1.2,\ 1.5\}$ for severity levels 1, 2, and 3.

\paragraph{Snow.}
Snow is implemented by overlaying randomly generated snowflakes together with a global contrast reduction. The snow intensity parameter is set to $\{0.4,\ 0.9,\ 1.1\}$ for severity levels 1, 2, and 3.

\paragraph{Camera Crash.}
Camera Crash is simulated by randomly selecting a subset of the six camera views and setting the corresponding images to the minimum value. The numbers of crashed views under the three severity levels are $\{1,\ 2,\ 4\}$.

\paragraph{Frame Lost.}
Frame Lost is simulated by randomly selecting a subset of the six camera views and replacing their current frames with frames sampled from the remaining non-lost views. The numbers of affected views under the three severity levels are $\{1,\ 2,\ 3\}$.

\subsection{Implementation Details of Adversarial Corruptions}

All adversarial attacks are applied directly to the multi-view camera inputs before BEV feature construction. 

FGSM~\citep{goodfellow2014explaining} is implemented as a single-step gradient-sign attack with perturbation budget 
$\epsilon \in \{1/255,\, 2/255,\, 4/255\}$.
In our implementation, FGSM optimizes a loss that combines binary cross-entropy with IoU-suppression terms in order to directly reduce BEV semantic prediction quality.

PGD~\citep{madry2017towards} is implemented as a multi-step projected attack with random uniform initialization in $[-\epsilon, \epsilon]$. 
The parameter settings under the three severity levels are as follows:
\begin{equation}
\begin{aligned}
(\epsilon, \alpha, \text{steps}) \in \{
&(1/255,\, 0.3/255,\, 5),\\
&(2/255,\, 0.5/255,\, 10),\\
&(4/255,\, 1/255,\, 15)
\}.
\end{aligned}
\end{equation}
In our implementation, PGD optimizes a loss that combines binary cross-entropy with IoU-suppression terms, and additionally includes a false-positive encouragement term on non-vehicle regions.

C\&W Attack~\citep{carlini2017towards} is implemented as an optimization-based attack with severity-dependent regularization strength, confidence target, iteration count, perturbation bound, and learning rate. 
The parameter settings under the three severity levels are as follows:
\begin{equation}
\begin{aligned}
(c, \kappa, \text{iter}, \epsilon, \text{lr}) \in \{
&(1.0,\, 0.15,\, 50,\, 4/255,\, 0.02),\\
&(3.0,\, 0.35,\, 80,\, 8/255,\, 0.02),\\
&(8.0,\, 0.55,\, 100,\, 16/255,\, 0.03)
\}.
\end{aligned}
\end{equation}
In our implementation, C\&W Attack minimizes an $L_2$-regularized objective that explicitly pushes the attacked prediction below an IoU target derived from the clean prediction.
\section{Implementation Details}
\label{sec:supp_impl}

\subsection{Inputs and Temporal Construction}
RESBev is designed to operate in the semantic feature space of an existing BEV perception pipeline, rather than directly on raw input images. In this work, RESBev is applied to several representative LSS-based BEV pipelines, including Lift-Splat-Shoot (LSS)~\citep{philion2020lift}, SimpleBEV~\citep{harley2022simple}, and BEVFormer~\citep{li2022bevformer}, and GaussianLSS~\citep{sonarghare2025fisheyegaussianlift}. 
At time step $t$, the current disturbed perception data is first processed by the original 2D image backbone and view transformer of the baseline model, producing a corrupted BEV feature map
\begin{equation}
f^{corrupt}_{t} \in \mathbb{R}^{C \times 200 \times 200}.
\end{equation}
In addition to the current corrupted feature, RESBev takes as input the reconstructed BEV feature from the previous step, and the corresponding ego-motion which encodes the relative translation and rotation between time steps $t-1$ and $t$.
\begin{equation}
f^{rec}_{t-1} \in \mathbb{R}^{C \times 200 \times 200}, \qquad
a_{t-1} \in \mathbb{R}^{15}.
\end{equation}

The full recurrent process is defined over a temporal horizon of length $\tau$ in the probabilistic formulation, while the actual recurrent prediction at inference time proceeds frame by frame. At each step, RESBev predicts a clean BEV prior from $(f^{rec}_{t-1}, a_{t-1})$, and then fuses this prior with the current corrupted observation $f^{corrupt}_{t}$ to produce the reconstructed feature $f^{rec}_{t}$.

\subsection{Overall Pipeline of RESBev}
RESBev consists of two core modules: a \emph{Semantic Prior Predictor} and an \emph{Anomaly Reconstructor}. The Semantic Prior Predictor is responsible for forecasting a clean semantic prior for the current BEV state from historical information and ego-motion. Taking the predicted prior as a robustness anchor, the Anomaly Reconstructor then selectively retrieves valid evidence from the current corrupted observation.

Formally, the overall pipeline can be summarized as:
\begin{equation}
f^{pred}_{t} = \mathcal{P}(f^{rec}_{t-1}, a_{t-1}), \qquad
f^{rec}_{t} = \mathcal{R}(f^{pred}_{t}, f^{corrupt}_{t}).
\end{equation}
where $\mathcal{P}(\cdot)$ denotes the Semantic Prior Predictor and $\mathcal{R}(\cdot)$ denotes the Anomaly Reconstructor.
The reconstructed feature $f^{rec}_{t}$ is finally fed into the task head of the baseline BEV perception model for semantic prediction.

\subsection{Implementation Details of Semantic Prior Predictor}
The Semantic Prior Predictor is implemented as a lightweight latent dynamics predictor composed of four parts: a visual encoder $E_{vis}$, an action encoder $E_{act}$, a Latent Dynamic Predictor (LDP), and a decoder $D$.

Given the previous reconstructed BEV feature $f^{rec}_{t-1}$ and the ego-motion vector $a_{t-1}$, we first map them into a compact latent space:
\begin{equation}
z^{vis}_{t-1} = E_{vis}(f^{rec}_{t-1}), \qquad
z^{act}_{t-1} = E_{act}(a_{t-1}).
\end{equation}
The visual latent is represented as a compact sequence of BEV latent tokens, while the action latent is represented as a single action token. The two latent representations are then concatenated into an action-aware latent state:
\begin{equation}
z_{t-1} = \mathrm{Concat}(z^{vis}_{t-1}, z^{act}_{t-1}).
\end{equation}
A latent dynamics predictor is then applied to predict the future latent state:
\begin{equation}
z^{pred}_{t} = \mathrm{LDP}(z_{t-1}),
\end{equation}
which is decoded back into the dense BEV feature space:
\begin{equation}
f^{pred}_{t} = D(z^{pred}_{t}).
\end{equation}
Combining the above steps gives the predictor formulation used in the main paper:
\begin{equation}
f^{pred}_{t}
=
D\Big(
\mathrm{LDP}\big(
\mathrm{Concat}(E_{vis}(f^{rec}_{t-1}), E_{act}(a_{t-1}))
\big)
\Big).
\end{equation}

\paragraph{Implementation details.}
For reproducibility, we report the detailed configuration of each component in the Semantic Prior Predictor. The visual encoder is a lightweight convolutional patch encoder composed of three strided $3\times3$ convolution blocks, followed by a linear projection that maps the encoded BEV feature into the latent space used by the predictor. The action encoder is a 2-layer MLP with architecture $15 \rightarrow 128 \rightarrow 256$ and GELU nonlinearity, which maps the ego-motion vector into an action token in the same latent space. The latent dimension is set to $d=256$, which is the shared embedding dimension for both visual latent tokens and the action token in the predictor. The number of visual latent tokens is $N=625$, corresponding to a $25 \times 25$ latent BEV grid obtained after spatial downsampling of the input BEV feature map. In addition, one action token produced by the action encoder is concatenated with the visual latent tokens before being fed into the latent dynamics predictor. The latent dynamics predictor(LDP) is implemented as a compact Transformer consisting of 4 encoder blocks with 8 attention heads and an MLP ratio of 4.0. The decoder is a lightweight BEV decoder that takes the predicted visual latent tokens, reshapes them back into a low-resolution latent BEV grid, and then progressively restores them to the original BEV feature resolution through two upsampling-convolution stages followed by final $3\times3$ convolutional refinement layers.

\begin{table}[b]
    \centering
    \caption{Representative feature flow in RESBev.}
    \label{tab:feature_shapes_impl}
    \begin{tabular}{lcc}
        \toprule
        Stage & Representation & Shape \\
        \midrule
        Previous reconstructed BEV & $f^{rec}_{t-1}$ & $C \times 200 \times 200$ \\
        Current corrupted BEV & $f^{corrupt}_{t}$ & $C \times 200 \times 200$ \\
        Ego-motion vector & $a_{t-1}$ & $15$ \\
        Visual latent & $z^{vis}_{t-1}$ & $625 \times 256$ \\
        Action latent & $z^{act}_{t-1}$ & $1 \times 256$ \\
        Predicted latent state & $z^{pred}_{t}$ & $625 \times 256$ \\
        Predicted BEV prior & $f^{pred}_{t}$ & $C \times 200 \times 200$ \\
        Projected query/key/value maps & $Q,K,V$ & $256 \times 200 \times 200$ \\
        Final reconstructed BEV & $f^{rec}_{t}$ & $C \times 200 \times 200$ \\
        \bottomrule
    \end{tabular}
\end{table}

\subsection{Implementation Details of Anomaly Reconstructor}
The Anomaly Reconstructor is a lightweight prior-guided fusion module.
Its role is not to independently generate a new BEV representation, but to selectively incorporate the extracted valid information from the current corrupted observation into the predicted clean prior.

Given the predicted BEV prior $f^{pred}_{t}$ and the current corrupted feature $f^{corrupt}_{t}$, we construct a prior-guided spatial cross-attention module directly on 2D BEV feature maps. In this module, the predicted prior serves as the query source, while the current corrupted feature provides the key-value context for selective feature retrieval:
\begin{equation}
Q = f^{pred}_{t}, \qquad
K = V = f^{corrupt}_{t}.
\end{equation}
All feature maps are first projected into a shared hidden space while preserving their 2D BEV structure. The reconstructed feature is then obtained through gated residual fusion:
\begin{equation}
f^{rec}_{t}
=
f^{pred}_{t}
+
\alpha \cdot \mathrm{CrossAttn}(Q, K, V),
\label{eq:supp_reconstructor}
\end{equation}
where $\alpha \in [0,1]$ is a learned dynamic gating factor.

\paragraph{Implementation details.}
For reproducibility, we report the detailed configuration of each component in the Anomaly Reconstructor. The prior-guided cross-attention is implemented directly on 2D BEV feature maps after linear projection into a shared hidden space, so that the predicted prior and the current corrupted feature can interact in a common representation space while preserving BEV spatial structure. The attention hidden dimension is set to 256, which is the shared channel dimension used for the projected query, key, and value features in the reconstructor. Multi-head attention is implemented with 8 attention heads. To preserve spatial correspondence in the BEV plane, 2D learnable positional embeddings are added to the projected feature maps before attention is applied. The dynamic fusion gate $\alpha$ is implemented as a lightweight 2-layer MLP with sigmoid activation. It takes the fused attention features as input and produces channel-wise gating coefficients, which control how much retrieved information is injected into the predicted BEV prior during reconstruction.

\subsection{Feature Shapes and Computational Flow}
We summarize the main feature flow of RESBev in Table~\ref{tab:feature_shapes_impl}.

\subsection{Optimization Details}
All models are implemented in PyTorch and trained on a single NVIDIA A100-SXM4-80GB GPU with a batch size of 16.

For reproducibility, the optimization details are summarized as follows:

\begin{itemize}
    \item optimizer: AdamW with $\beta_1=0.9$, $\beta_2=0.999$, and $\epsilon=10^{-8}$;
    \item initial learning rate: $5\times10^{-4}$;
    \item weight decay: $1\times10^{-4}$;
    \item scheduler and warmup: linear warmup for the first 5 epochs starting from 0.2 times the base learning rate, followed by cosine decay; an auxiliary ReduceLROnPlateau scheduler is additionally used with factor 0.5, patience 5, and minimum learning rate $2\times10^{-5}$;
    \item total training epochs / iterations: 100 epochs;
    \item gradient clipping / mixed precision: gradient clipping with a maximum norm of 0.5; mixed precision is not used during training unless otherwise stated. Efficiency measurements are conducted under FP16 inference following the main paper.
\end{itemize}

\begin{table*}[tb]
    \centering
    \caption{Performance (IoU) of baseline models with and without RESBev across seen corruptions.}
    \label{tab:main_results_table_fixed}
    \setlength{\tabcolsep}{4pt}
    \resizebox{0.8\textwidth}{!}{%
    \begin{tabular}{lc c c c c c c c c}
        \toprule
        \multirow{2}{*}{\textbf{Corruption}} 
        & \multirow{2}{*}{\textbf{Severity}} 
        & \multicolumn{2}{c}{\textbf{LSS}} 
        & \multicolumn{2}{c}{\textbf{SimpleBEV}} 
        & \multicolumn{2}{c}{\textbf{GaussianLSS}} 
        & \multicolumn{2}{c}{\textbf{BEVFormer}} \\
        \cmidrule(lr){3-4} 
        \cmidrule(lr){5-6} 
        \cmidrule(lr){7-8} 
        \cmidrule(lr){9-10}
        & & Vanilla & +RESBev 
          & Vanilla & +RESBev 
          & Vanilla & +RESBev 
          & Vanilla & +RESBev \\
        \midrule
        \textbf{Clean} & - & 33.03 & 33.31 & 55.70 & 55.91 & 42.80 & 43.14 & 46.70 & 47.01 \\
        \midrule

        & \textbf{1} & 13.87 & 28.58 & 17.65 & 31.03 & 11.76 & 31.49 & 29.10 & 38.86 \\
        & \textbf{2} & 10.06 & 28.47 & 12.51 & 31.13 & 9.43 & 31.42 & 21.51 & 38.79 \\
        \textbf{FGSM} & \textbf{3} & 6.92 & 28.21 & 9.54 & 31.12 & 8.53 & 31.37 & 14.01 & 38.69 \\
        \cmidrule(lr){2-10}
        & \textbf{Avg.} & 10.28 & 28.42 & 13.23 & 31.09 & 9.91 & 31.43 & 21.54 & 38.78 \\
        \midrule

        & \textbf{1} & 12.55 & 31.96 & 15.44 & 30.84 & 10.23 & 31.21 & 26.08 & 38.95 \\
        & \textbf{2} & 8.93 & 31.34 & 11.25 & 31.68 & 7.64 & 31.04 & 18.57 & 38.88 \\
        \textbf{PGD} & \textbf{3} & 6.04 & 31.12 & 8.97 & 32.42 & 5.28 & 31.02 & 11.39 & 38.87 \\
        \cmidrule(lr){2-10}
        & \textbf{Avg.} & 9.17 & 31.47 & 11.89 & 31.65 & 7.72 & 31.09 & 18.68 & 38.90 \\
        \midrule

        & \textbf{1} & 13.91 & 28.46 & 22.42 & 33.83 & 27.12 & 32.09 & 32.02 & 39.00 \\
        & \textbf{2} & 9.24 & 28.38 & 16.76 & 33.81 & 25.98 & 31.94 & 25.31 & 38.93 \\
        \textbf{Fog} & \textbf{3} & 6.64 & 28.33 & 12.37 & 33.64 & 19.67 & 31.82 & 18.69 & 38.80 \\
        \cmidrule(lr){2-10}
        & \textbf{Avg.} & 9.93 & 28.39 & 17.18 & 33.76 & 24.26 & 31.95 & 25.34 & 38.91 \\
        \midrule

        & \textbf{1} & 13.32 & 28.44 & 21.39 & 33.74 & 27.10 & 32.05 & 29.31 & 38.78 \\
        & \textbf{2} & 9.35 & 28.37 & 17.56 & 33.91 & 25.60 & 31.87 & 25.64 & 38.73 \\
        \textbf{Bright} & \textbf{3} & 6.66 & 28.32 & 13.94 & 33.72 & 19.38 & 31.70 & 22.15 & 38.71 \\
        \cmidrule(lr){2-10}
        & \textbf{Avg.} & 9.78 & 28.38 & 17.63 & 33.79 & 24.03 & 31.87 & 25.70 & 38.74 \\
        \midrule

        & \textbf{1} & 15.24 & 28.41 & 23.54 & 33.85 & 22.23 & 31.85 & 30.47 & 38.84 \\
        & \textbf{2} & 9.43 & 28.32 & 16.11 & 33.64 & 12.76 & 31.53 & 25.81 & 38.76 \\
        \textbf{Frame Lost} & \textbf{3} & 7.28 & 28.27 & 13.58 & 33.70 & 9.12 & 31.44 & 21.30 & 38.65 \\
        \cmidrule(lr){2-10}
        & \textbf{Avg.} & 10.65 & 28.33 & 17.74 & 33.73 & 14.70 & 31.61 & 25.86 & 38.75 \\
        \bottomrule
    \end{tabular}%
    }
\end{table*}


\section{Additional Experiments}
\label{sec:supp_exp}

\subsection{Per-Severity Results on Seen Corruptions}
We provide the detailed per-severity results on the seen corruption set in Table~\ref{tab:main_results_table_fixed}. These results complement the main paper, which only reports the average across the three severity levels.

\begin{table*}[tb]
    \centering
    \caption{Performance (IoU) of baseline models with and without RESBev across unseen corruptions.}
    \label{tab:suppl_additional_corruptions}
    \setlength{\tabcolsep}{4pt}
    \resizebox{0.8\textwidth}{!}{%
    \begin{tabular}{lc c c c c c c c c}
        \toprule
        \multirow{2}{*}{\textbf{Corruption}} 
        & \multirow{2}{*}{\textbf{Severity}} 
        & \multicolumn{2}{c}{\textbf{LSS}} 
        & \multicolumn{2}{c}{\textbf{SimpleBEV}} 
        & \multicolumn{2}{c}{\textbf{GaussianLSS}} 
        & \multicolumn{2}{c}{\textbf{BEVFormer}} \\
        \cmidrule(lr){3-4} 
        \cmidrule(lr){5-6} 
        \cmidrule(lr){7-8} 
        \cmidrule(lr){9-10}
        & & Vanilla & +RESBev 
          & Vanilla & +RESBev 
          & Vanilla & +RESBev 
          & Vanilla & +RESBev \\
        \midrule
        \textbf{Clean} & - & 33.03 & 33.31 & 55.70 & 55.91 & 42.80 & 43.14 & 46.70 & 47.01 \\
        \midrule

        & \textbf{1} & 11.28 & 30.94 & 14.65 & 30.87 & 8.02 & 31.38 & 24.96 & 39.71 \\
        & \textbf{2} & 8.54 & 30.75 & 11.67 & 30.74 & 6.09 & 31.21 & 17.54 & 39.70 \\
        \textbf{C\&W Attack} & \textbf{3} & 6.51 & 30.70 & 7.98 & 30.22 & 3.80 & 31.12 & 10.18 & 39.60 \\
        \cmidrule(lr){2-10}
        & \textbf{Avg.} & 8.78 & 30.80 & 11.43 & 30.61 & 5.97 & 31.24 & 17.56 & 39.67 \\
        \midrule

        & \textbf{1} & 17.69 & 28.44 & 22.85 & 33.67 & 21.23 & 32.16 & 32.85 & 38.95 \\
        & \textbf{2} & 8.34 & 28.34 & 16.76 & 33.69 & 16.56 & 32.09 & 27.09 & 38.93 \\
        \textbf{Snow} & \textbf{3} & 4.76 & 28.27 & 12.59 & 33.78 & 10.45 & 32.05 & 24.39 & 38.85 \\
        \cmidrule(lr){2-10}
        & \textbf{Avg.} & 10.26 & 28.35 & 17.40 & 33.71 & 16.08 & 32.10 & 28.11 & 38.91 \\
        \midrule

        & \textbf{1} & 14.21 & 28.39 & 23.88 & 33.71 & 19.53 & 31.73 & 31.05 & 39.79 \\
        & \textbf{2} & 10.38 & 28.33 & 16.90 & 33.74 & 13.47 & 31.57 & 27.42 & 39.74 \\
        \textbf{Camera Crash} & \textbf{3} & 5.64 & 28.31 & 11.24 & 33.76 & 7.24 & 31.38 & 23.91 & 39.63 \\
        \cmidrule(lr){2-10}
        & \textbf{Avg.} & 10.08 & 28.34 & 17.34 & 33.74 & 13.41 & 31.56 & 27.46 & 39.72 \\
        \midrule

        & \textbf{1} & 10.49 & 28.36 & 21.67 & 33.71 & 24.03 & 31.80 & 28.96 & 39.43 \\
        & \textbf{2} & 8.51 & 28.25 & 17.56 & 33.86 & 15.85 & 31.44 & 20.51 & 39.35 \\
        \textbf{Noise} & \textbf{3} & 6.93 & 28.21 & 13.57 & 33.71 & 10.12 & 31.04 & 15.18 & 39.24 \\
        \cmidrule(lr){2-10}
        & \textbf{Avg.} & 8.64 & 28.27 & 17.60 & 33.76 & 16.67 & 31.43 & 21.55 & 39.34 \\
        \midrule

        & \textbf{1} & 11.76 & 28.39 & 20.98 & 33.73 & 25.04 & 32.02 & 31.99 & 39.82 \\
        & \textbf{2} & 7.19 & 28.35 & 16.32 & 33.77 & 16.98 & 31.95 & 27.47 & 39.73 \\
        \textbf{Dark} & \textbf{3} & 5.37 & 28.33 & 11.56 & 33.75 & 11.02 & 31.91 & 24.72 & 39.70 \\
        \cmidrule(lr){2-10}
        & \textbf{Avg.} & 8.11 & 28.36 & 16.29 & 33.75 & 17.68 & 31.96 & 28.06 & 39.75 \\
        \bottomrule
    \end{tabular}%
    }
\end{table*}

\subsection{Per-Severity Results on Unseen Corruptions}
We provide the detailed per-severity results on unseen corruptions in Table~\ref{tab:suppl_additional_corruptions}.

\subsection{Long-Horizon Robustness on Unseen Corruptions}

We further evaluate RESBev under consecutive unseen corruptions using the same 50-step recursive reconstruction setting as in the main paper. 
As shown in Table~\ref{tab:supp_consecutive_unseen_corruptions}, RESBev maintains stable performance under long corrupted sequences, indicating that the recurrent predictive prior generalizes beyond the corruptions observed during training.

\begin{table}[t]
    \centering
    \small
    \caption{
    Long-horizon robustness under consecutive unseen corruptions.
    }
    \label{tab:supp_consecutive_unseen_corruptions}
    \setlength{\tabcolsep}{3.5pt}
    \resizebox{0.7\textwidth}{!}{%
    \begin{tabular}{l c c c c >{\columncolor{gray!15}}c}
        \toprule
        \textbf{Corruption} & \textbf{1-Step} & \textbf{10-Step} & \textbf{30-Step} & \textbf{50-Step} & \textbf{Change} \\
        \midrule
        C\&W & 30.80 & 30.26 & 30.30 & 30.13 & $\downarrow$2.18\% \\
        Snow & 28.35 & 27.86 & 27.81 & 27.80 & $\downarrow$1.94\% \\
        Camera Crash & 28.34 & 28.00 & 27.91 & 27.79 & $\downarrow$1.94\% \\
        Noise & 28.27 & 27.98 & 27.81 & 27.80 & $\downarrow$1.66\% \\
        Dark & 28.36 & 27.93 & 27.95 & 27.88 & $\downarrow$1.69\% \\
        \bottomrule
    \end{tabular}%
    }
\end{table}

\section{Discussions}
\label{sec:supp_discussion}

\subsection{Comparison with Existing Robustness Strategies}

Compared with fusion-based BEV perception methods such as BEVCar~\citep{schramm2024bevcar}, RESBev does not rely on additional sensing modalities and is therefore easier to integrate into camera-only BEV perception systems. This makes RESBev especially attractive in practical settings where sensor cost, deployment complexity, or hardware constraints must be carefully controlled. In contrast, fusion-based methods can leverage complementary physical observations across modalities and may retain stronger robustness when the visual stream is severely degraded.

Compared with temporal aggregation methods such as BEVFormer~\citep{li2022bevformer}, RESBev explicitly predicts a clean semantic prior rather than merely aggregating historical and current observations. This design makes it better suited to bypass corrupted inputs and recover temporally consistent semantic structure under heterogeneous corruptions. By introducing a predictive prior before feature reconstruction, RESBev emphasizes semantic recovery instead of relying solely on direct temporal fusion.

Compared with graph-based reasoning methods such as GraphBEV~\citep{song2024graphbev}, RESBev provides a predictive reconstruction perspective that is naturally applicable to both natural corruptions and adversarial perturbations. While graph-based approaches are particularly effective at modeling structured relations and long-range interactions, RESBev focuses on recovering corrupted BEV representations through temporally grounded semantic prediction, offering a complementary robustness mechanism.

Compared with diffusion-based BEV refinement methods such as DiffBEV~\citep{zou2024diffbev} and BEVDiffuser~\citep{ye2025bevdiffuser}, RESBev avoids iterative diffusion sampling and instead performs one-step temporal prior prediction followed by prior-guided reconstruction. Diffusion-based methods are effective at denoising or refining corrupted BEV features, but they typically introduce additional inference cost due to their generative refinement process. In contrast, RESBev uses historical BEV dynamics and ego-motion to predict a semantic prior, enabling lightweight feature recovery with limited deployment overhead. These two directions are complementary: diffusion-based refinement emphasizes generative denoising, while RESBev emphasizes temporal semantic prediction and selective reconstruction.

\subsection{Future Directions}

Several directions may further strengthen the proposed framework. First, although RESBev already shows effectiveness on both BEV semantic segmentation and 3D object detection, it is still valuable to extend the framework to a broader range of BEV perception tasks, such as occupancy prediction and online mapping. Such extensions would help clarify whether predictive semantic reconstruction provides equally strong benefits across different BEV outputs and downstream objectives.

Second, it is worthwhile to further investigate how the proposed predictive prior can be incorporated into more transformer-based BEV architectures. Since transformer-based models often rely on BEV queries and spatiotemporal attention for feature construction and temporal interaction, establishing a compatible predictive reconstruction mechanism in this setting would further broaden the applicability of RESBev beyond the current evaluated backbones.

Third, future work may explore stronger uncertainty modeling, multi-hypothesis state prediction, and more expressive dynamic scene modeling in order to better handle highly stochastic traffic scenarios and long-horizon corruptions. These directions may further improve the reliability and generalizability of predictive robustness for real-world BEV perception systems.

Finally, it would also be valuable to study whether RESBev can be combined with complementary robustness strategies, including diffusion-based refinement, stronger temporal modeling, structured relational reasoning, or multi-modal perception, so as to further improve robustness under more challenging real-world anomalies.


\newpage

\end{document}